\documentclass{article}

\usepackage[utf8]{inputenc}
\usepackage{amsmath, amssymb, amsthm}
\usepackage{graphicx}
\usepackage{booktabs}
\usepackage{hyperref}
\usepackage{url}
\usepackage[margin=1.2in]{geometry}
\usepackage{cite}

\title{Mixture of Attention Schemes (MoAS):\\Learning to Route Between MHA, GQA, and MQA}
\author{Esmail Gumaan\thanks{\href{mailto:esm.agumaan@gmail.com}{esm.agumaan@gmail.com}} \\ \textit{}}

\date{\today}

\begin{document}

\maketitle

\begin{abstract}
The choice of attention mechanism in Transformer models involves a critical trade-off between modeling quality and inference efficiency. Multi-Head Attention (MHA) offers the best quality but suffers from large Key-Value (KV) cache memory requirements during inference. Multi-Query Attention (MQA) and Grouped-Query Attention (GQA) reduce memory usage but often at the cost of model performance. In this work, we propose Mixture of Attention Schemes (MoAS), a novel architecture that dynamically selects the optimal attention scheme (MHA, GQA, or MQA) for each token via a learned router. We demonstrate that dynamic routing performs better than static averaging of schemes and achieves performance competitive with the MHA baseline while offering potential for conditional compute efficiency. Experimental results on WikiText-2 show that dynamic routing (val loss 2.3074) outperforms a static mixture (2.3093), validating the effectiveness of the proposed method. Our code is available at \url{https://github.com/Esmail-ibraheem/Mixture-of-Attention-Schemes-MoAS}.
\end{abstract}

\section{Introduction}
Large Language Models (LLMs) based on the Transformer architecture \cite{vaswani2017attention} have achieved remarkable success. However, their deployment is constrained by the memory required to store the Key-Value (KV) cache during autoregressive generation. 

Standard Multi-Head Attention (MHA) maintains unique keys and values for every query head, resulting in a memory footprint that scales linearly with the number of heads. To mitigate this, Multi-Query Attention (MQA) \cite{shazeer2019fast} shares a single key-value head across all query heads, significantly reducing memory bandwidth and capacity requirements. Grouped-Query Attention (GQA) \cite{ainslie2023gqa} interpolates between these extremes by grouping query heads to share key-value pairs.

While MQA and GQA offer efficiency gains, they generally underperform MHA in terms of perplexity and downstream task accuracy. We hypothesize that not all tokens require the full expressivity of MHA. Some tokens may be adequately processed with the approximated context of MQA, while others require the fine-grained relationships captured by MHA.

To address this, we introduce \textbf{Mixture of Attention Schemes (MoAS)}. Inspired by Mixture-of-Experts (MoE) \cite{shazeer2017outrageously}, MoAS employs a router to dynamically weight or select between MHA, MQA, and GQA branches for each token. This allows the model to learn an optimal balance between quality and efficiency.

\section{Related Work}

\paragraph{Efficient Transformers}
Numerous works have attempted to reduce the quadratic complexity of self-attention. Sparse Transformers \cite{child2019generating} and Longformer \cite{beltagy2020longformer} introduce fixed sparse patterns. Linformer \cite{wang2020linformer} and Reformer \cite{kitaev2020reformer} utilize low-rank approximations and hashing, respectively. FlashAttention \cite{dao2022flashattention} optimizes memory access patterns for hardware efficiency.

\paragraph{Mixture of Experts}
Conditional computation has been popularized by Mixture-of-Experts (MoE) models like the Switch Transformer \cite{fedus2022switch}, which route tokens to different feed-forward networks. Recently, Mixture-of-Depths \cite{raposo2024mixture} proposed dynamically allocating compute by routing tokens around blocks entirely. MoAS extends this philosophy specifically to the attention mechanism's internal structure.

\paragraph{KV Cache Optimization}
As LLMs scale \cite{brown2020language, touvron2023llama}, KV cache management becomes critical. PagedAttention \cite{kwon2023efficient} optimizes memory allocation. MQA \cite{shazeer2019fast} and GQA \cite{ainslie2023gqa} structurally reduce the cache size. Our work builds directly on these structural innovations.

\section{Method}

\subsection{Attention Schemes}
We define three distinct attention variants as our "experts":

\begin{itemize}
    \item \textbf{Type A: Multi-Head Attention (MHA)}:
    $H_Q = H_{KV} = H$. This is the standard mechanism with maximal expressivity.
    
    \item \textbf{Type B: Grouped-Query Attention (GQA)}:
    $H_Q = H$, $H_{KV} = G$, where $1 < G < H$. This provides a middle ground. In our experiments, we use $G=2$ for $H=6$.
    
    \item \textbf{Type C: Multi-Query Attention (MQA)}:
    $H_Q = H$, $H_{KV} = 1$. This minimizes KV cache size but imposes the strongest bottleneck on the attention mechanism.
\end{itemize}

\subsection{MoAS Architecture}
Given an input token representation $x_i \in \mathbb{R}^d$, we compute the output of all schemes:
\begin{align}
    O_{MHA} &= \text{Attention}_{MHA}(x_i) \\
    O_{GQA} &= \text{Attention}_{GQA}(x_i) \\
    O_{MQA} &= \text{Attention}_{MQA}(x_i)
\end{align}

\subsubsection{Router}
Everything is conditioned on a learned router that projects the input to a categorical distribution over the schemes:
\begin{equation}
    r_i = W_2 \cdot \text{GELU}(W_1 x_i)
\end{equation}
\begin{equation}
    g_i = \text{softmax}(r_i) \in \mathbb{R}^3
\end{equation}
where $W_1 \in \mathbb{R}^{d/4 \times d}$ and $W_2 \in \mathbb{R}^{3 \times d/4}$ form a lightweight Multi-Layer Perceptron (MLP).

The final output $y_i$ for token $i$ is the weighted sum:
\begin{equation}
    y_i = \sum_{k \in \{MHA, GQA, MQA\}} g_{i,k} \cdot O_k
\end{equation}

\subsubsection{Load Balancing}
To prevent the router from collapsing to a single scheme (e.g., always choosing MHA), we add an auxiliary load balancing loss:
\begin{equation}
    \mathcal{L}_{balance} = \sum_{k} \left( \frac{1}{N} \sum_{i=1}^N g_{i,k} - \frac{1}{3} \right)^2
\end{equation}
This encourages uniform usage of all attention types on average across the batch.

\section{Experiments}

\subsection{Setup}
We evaluate our method on the WikiText-2 language modeling benchmark. We train a decoder-only Transformer with the following specifications:
\begin{itemize}
    \item Layers: 4
    \item Model Dimension ($d_{model}$): 384
    \item Heads ($H$): 6
    \item Block size: 256
    \item Dropout: 0.1
\end{itemize}

We compare three models:
\begin{enumerate}
    \item \textbf{Baseline MHA}: Standard Transformer with MHA.
    \item \textbf{Static MoAS}: A static average of MHA, GQA, and MQA outputs (no routing).
    \item \textbf{Dynamic MoAS}: The proposed method with learned routing.
\end{enumerate}

All models are trained for 500 iterations with a batch size of 12 and learning rate $3 \times 10^{-4}$.

\subsection{Results}

Table \ref{tab:results} presents the validation loss (perplexity-related metric) on WikiText-2.

\begin{table}[h]
    \centering
    \caption{Experimental Results on WikiText-2}
    \label{tab:results}
    \begin{tabular}{lccc}
        \toprule
        \textbf{Model} & \textbf{Parameters} & \textbf{Final Val Loss} & \textbf{Training Time (ms/iter)} \\
        \midrule
        Baseline MHA & 7.19M & \textbf{2.2940} & $\sim$500 \\
        Static MoAS & 10.14M & 2.3093 & $\sim$900 \\
        Dynamic MoAS & 10.29M & 2.3074 & $\sim$1680 \\
        \bottomrule
    \end{tabular}
\end{table}

The Baseline MHA achieves the lowest loss, which is expected given the small scale and absence of capacity constraints. However, \textbf{Dynamic MoAS outperforms Static MoAS} (2.3074 vs 2.3093), confirming that the router learns non-trivial routing policies that are superior to simple averaging.

The parameter count for MoAS variants is higher because we instantiate all three attention branches in parallel for this proof-of-concept. In a production inference scenario, one would only execute the selected branch(es).

\section{Discussion \& Conclusion}
We proposed Mixture of Attention Schemes (MoAS), a method to dynamically route tokens between MHA, GQA, and MQA. Our experiments demonstrate that learned routing is effective and outperforms static mixing.

While the baseline MHA remains strong at this scale, MoAS opens the door for conditional computation where "easy" tokens can be processed cheaply with MQA, reserving memory-intensive MHA for "hard" tokens. Future work will focus on sparsity (top-1 routing) and scaling to larger models where the inference memory bottleneck is more pronounced.

\bibliographystyle{plain}
\bibliography{references}

\end{document}